\documentclass{article} 
\usepackage{iclr2027_conference,times}

\usepackage{amsmath,amsfonts,bm}

\def\eqref#1{equation~\ref{#1}}

\def\1{\bm{1}}

\DeclareMathAlphabet{\mathsfit}{\encodingdefault}{\sfdefault}{m}{sl}
\SetMathAlphabet{\mathsfit}{bold}{\encodingdefault}{\sfdefault}{bx}{n}

\usepackage{hyperref}
\usepackage{url}
\usepackage{graphicx}
\usepackage{makecell}
\usepackage{amsmath,amssymb, algorithm,algpseudocode}
\usepackage{multirow}
\usepackage[table]{xcolor}
\usepackage[xcdraw]{xcolor}
\usepackage{booktabs}
\usepackage[normalem]{ulem}
\useunder{\uline}{\ul}{}

\title{Learning from Failure: Leveraging Unreliable Predictions in Semi-Supervised Real-World Adverse Weather Removal}

\author{Cap Dang Xuan Kiet, Tat-Jen Cham \\
College of Computing and Data Science, Nanyang Technological University, Singapore\\
\texttt{capd0001@e.ntu.edu.sg}, \texttt{ASTJCham@ntu.edu.sg} \\
}

\iclrfinalcopy 
\begin{document}

\maketitle

\begin{abstract}
Adverse weather image restoration aims to recover images degraded by rain, haze, snow, and other weather-induced artifacts, thereby improving the robustness of outdoor vision systems. Existing unified restoration models exhibit limited generalization to real-world scenes due to their reliance on synthetic supervision and insufficient semantic constraints. In this paper, we propose a novel student--teacher semi-supervised framework that addresses both challenges. Specifically, we introduce an unreliable database that preserves failed teacher predictions as informative negative samples for contrastive learning, while a reliable database stores high-quality teacher predictions as positive samples. By jointly exploiting reliable pseudo-ground truths and unreliable teacher outputs, the proposed framework learns to enhance desirable restoration characteristics while avoiding common failures. We further propose a phase spectrum-based semantic constraint that replaces computationally expensive text-based supervision with an efficient and naturally aligned semantic prior. An adaptive phase consistency loss is also designed to dynamically balance supervision between the degraded input and teacher pseudo-ground truths according to degradation severity. Extensive experiments on real-world benchmarks demonstrate that the proposed method consistently outperforms existing state-of-the-art approaches in restoration quality and perceptual fidelity while exhibiting stronger generalization to real-world adverse weather conditions.

\end{abstract}


\section{Introduction}

Images captured in real-world environments are often degraded by adverse weather conditions such as rain, haze, and snow, which significantly deteriorate the performance of outdoor vision systems. Existing image restoration methods have mainly focused on task-specific models tailored to individual weather phenomena \citep{Wang2025LearningHT}. Although these methods achieve strong performance through extensive training in large-scale synthetic datasets, their highly specialized designs limit scalability in diverse weather conditions. 

To address this limitation, recent studies have proposed unified restoration frameworks capable of handling multiple weather degradations within a single model \citep{Wang2025MODEMAM,Wu2025GradientAC}. Although these approaches demonstrate promising results on synthetic benchmarks, they exhibit limited generalization to real-world scene. We attribute this to two key factors: their heavy reliance on synthetic supervision and neglect of the semantic context of the scenes, due to insufficient constraints during training.

To alleviate the dependency on synthetic data, unsupervised restoration methods have been proposed that require only unpaired real-world images \citep{Liu2025FrequencyDD}. However, these approaches fail to exploit the abundant labeled synthetic datasets that have become widely available. Semi-supervised learning offers a more attractive compromise by leveraging both labeled synthetic and unlabeled real-world data. For example, WResVLM \citep{Xu2024TowardsRA} adopts a student--teacher framework in which the teacher generates pseudo-ground truths (pseudo-GTs) for unlabeled images. Nevertheless, the framework performs contrastive learning using only positive samples derived from reliable pseudo-ground truths, overlooking informative negative samples that could explicitly guide the student away from poor restoration results. We observe that the teacher frequently produces low-quality predictions during training. These predictions often contain severe artifacts, color distortions, or incomplete restoration (Fig.~\ref{fig:test_with_teacher}b), and have previously been simply discarded. However, we note that they are \emph{not just} generic negative samples, but are in fact sampled directly from the model in an \emph{on-policy} manner, and hence provide valuable information \citep{tang2024understandingperformancegaponline}. To address the previous limitation, we propose a mechanism that leverages these informative failures for contrastive semi-supervised learning. More specifically, we retain these cases in an unreliable database and treat them as negative samples for contrastive learning. By simultaneously attracting the student toward reliable restorations while repelling it from unreliable ones, the proposed framework provides stronger supervision and improves generalization to real-world adverse weather conditions.

Besides the supervision strategy, effective semantic guidance is equally important for restoring visually plausible images. Recent methods employ large vision-language models (VLMs) to generate textual descriptions that explicitly constrain image semantics \citep{Xu2024TowardsRA}. Although effective, text supervision suffers from two inherent drawbacks. First, textual descriptions are only weakly aligned with the fine-grained spatial structures required for low-level vision tasks because of language ambiguity and the lossy language-vision representation gap \citep{Wu2023QBenchAB}. Second, generating accurate descriptions with large VLMs introduces substantial computational overhead,  slowing the training process.

Instead of relying on language supervision, we exploit the Fourier phase spectrum as an efficient structural prior. It is well established that the Fourier phase spectrum primarily preserves structural and semantic information, whereas the amplitude spectrum mainly represents low-level appearance \citep{Oppenheim1979PhaseIS}. Existing Fourier-based restoration methods \citep{Guo2022ExploringFP} mainly focus on recovering degraded amplitude information. In contrast, we employ the phase spectrum to constrain structural consistency during semi-supervised learning. Even though the pseudo-GTs are accessible, using them alone is not optimal. \citep{Oppenheim1980TheIO} shows that the phase spectrum is inherently robust to illumination variations and color shifts, making it possible to also use the phase spectrum of the input image to constrain the structure of the output image. Nevertheless, severe weather degradations may also corrupt phase information by obscuring structural details \citep{Liu2025FrequencyDD}. Therefore, directly enforcing consistency with the phase spectrum of either the degraded input or pseudo-GT can lead to sub-optimal solutions. To overcome this issue, we propose an \emph{adaptive phase consistency loss} that dynamically balances supervision between the degraded input and the teacher-generated pseudo-GT according to degradation severity. Consequently, the model relies more on the input phase for mild degradations while progressively shifting toward the pseudo-GT phase under severe degradations.

Our contributions are summarized as follows:
\begin{itemize}  
\item We propose a new mechanism in the family of semi-supervised weather removal methods that identifies and preserves failed teacher predictions in an unreliable database, enabling their use as negative samples for contrastive semi-supervised learning and improving generalization to real-world adverse weather image restoration.
\item We introduce the Fourier phase spectrum as an efficient structural prior, eliminating the need for computationally expensive and ambiguous text-based supervision.
\item We propose an adaptive phase consistency loss that automatically adjusts structural supervision according to degradation severity, producing more robust restorations across diverse adverse weather conditions.
\item The proposed framework establishes new state-of-the-art performance among unified and semi-supervised image restoration methods, outperforming existing approaches on both quantitative and qualitative evaluation.
\end{itemize}

    

\section{Related Works}
\subsection{Task-specific Image Restoration}
Image restoration is a fundamental problem in computer vision, aiming to recover high-quality images from degraded observations. Early task-specific methods \citep{Jiang2017ANT,Ancuti2013SingleID} primarily relied on physical imaging models and handcrafted priors tailored to individual degradation types. With the advent of convolutional neural networks \citep{Yang2024DepthAwareUV,Yang2023RobustUI} and, more recently, transformer-based architectures have significantly advanced restoration performance across a wide range of tasks. For example, Restormer \citep{Zamir2021RestormerET} introduces an efficient multi-scale attention mechanism that achieves an excellent balance between restoration quality and computational efficiency. These transformer-based architectures have been successfully applied to deraining \citep{Chen2023LearningAS}, desnowing \citep{Zhang2022DesnowFormerAE}, and dehazing \citep{song2023vision}. Despite their impressive performance, most existing methods remain task-specific and require separate models for different degradation types, leading to increased training, storage, and deployment costs.

\subsection{All-in-One Image Restoration}
To mitigate the inefficiency of deploying multiple models for diverse weather degradations, unified image restoration methods have attracted increasing attention \citep{zhu2023Weather,Ye_2023_ICCV}. While these approaches achieve impressive performance on synthetic benchmarks, they often struggle in real-world scenarios due to the substantial domain gap between synthetic and real-world degradations. To bridge this gap, \citep{Liu2025RealWorldAW} proposed a self-evolving framework that combines physics-driven synthetic data generation with a dual-level reinforcement learning strategy. Meanwhile, \citep{11460995} addressed the problem in an unsupervised setting employing global and local discriminators to enhance image realism while preserving structural consistency. WResVLM \citep{Xu2024TowardsRA} introduced a semi-supervised framework that leverages both labeled synthetic data and unlabeled real-world data through VLM-based semantic supervision. Although effective, the reliance on large VLMs incurs substantial computational overhead and limits scalability and adaptability to resource-constrained applications. we observe that the Fourier phase spectrum inherently preserves the structural and semantic information of an image while being significantly more computationally efficient than VLM-based representations. Motivated by this observation, we propose a phase spectrum-based semantic constraint that provides explicit semantic guidance for semi-supervised learning

\section{Methodology} \label{sec:method}
\begin{figure*}[t]
  \centering

    \includegraphics[width=\linewidth]{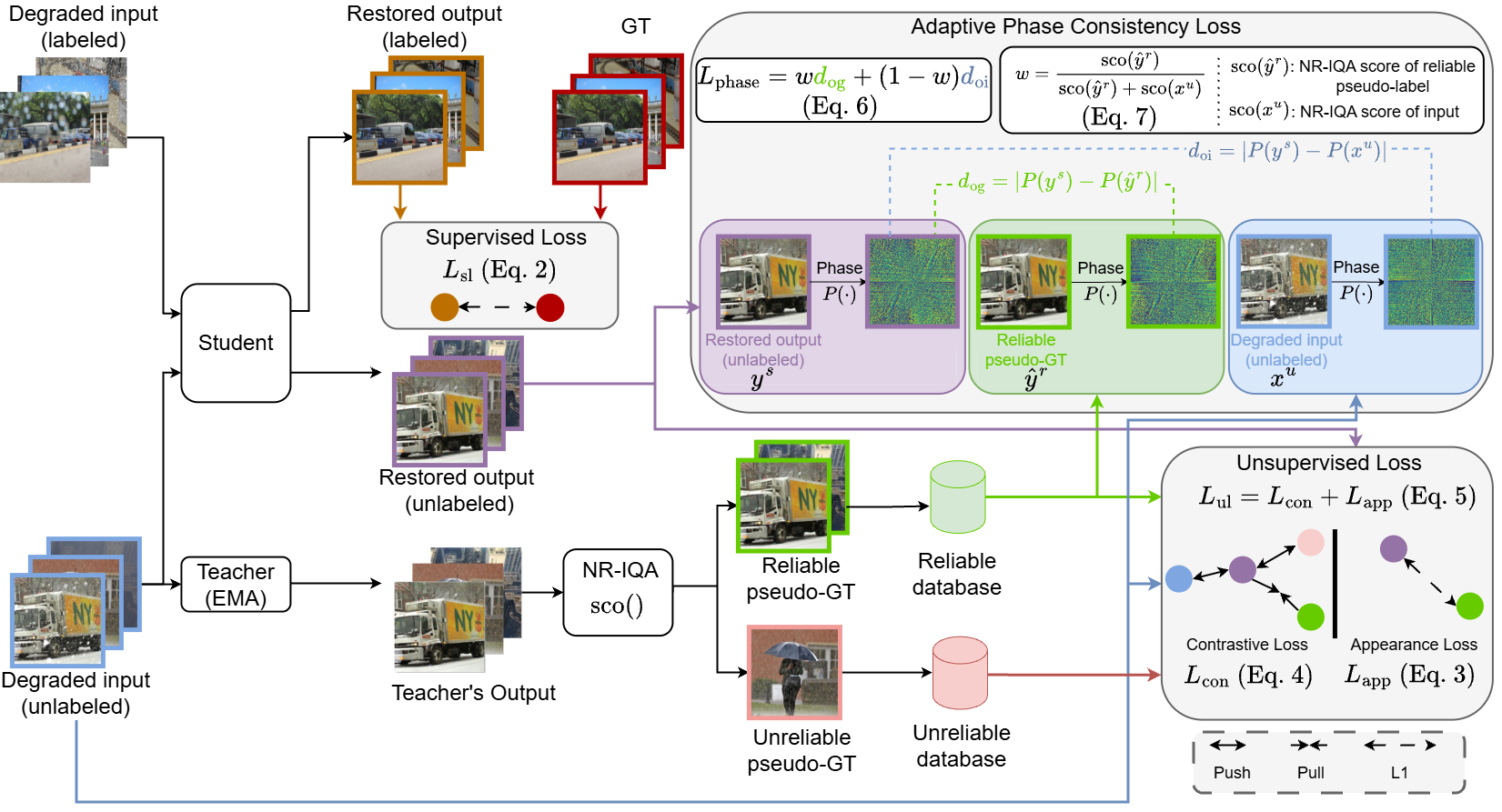}   
    \caption{The illustration of the proposed semi-supervised learning framework.}
  
  \label{fig:architecture}
\end{figure*}
\subsection{Overall Framework}
Motivated by the observations that existing semi-supervised image restoration methods lack informative negative samples and that VLM-based semantic constraints are computationally expensive, we propose a novel semi-supervised framework that addresses both limitations. Specifically, we incorporate an unreliable image database into the unsupervised learning process to provide challenging negative samples, while replacing conventional VLM-based semantic supervision with a lightweight phase spectrum-based semantic constraint. This design improves both training effectiveness and computational efficiency.

As illustrated in Fig.~\ref{fig:architecture}, our method follows the conventional student--teacher paradigm widely adopted in semi-supervised learning \citep{Sohn2020FixMatchSS}. The framework consists of a student network $f_{\theta_s}$ and a teacher network $f_{\theta_t}$ sharing the same architecture. The teacher parameters are updated using the exponential moving average (EMA) of the student parameters, while the student network is optimized by minimizing the following objective,
\begin{equation}
    L_\text{total} = L_\text{sl} + \alpha L_\text{ul} + \beta L_\text{phase}\>,
\end{equation}
where 
\begin{equation}
    L_\text{sl}=\sum_{i=1}^{N}|f_{\theta_s}(x_i^{l})-y_i^{l}|
\end{equation}
denotes the supervised reconstruction loss on the labeled dataset, $L_{ul}$ denotes the unsupervised loss (Eq.~\ref{eq:unsup_loss}), and $L_\text{phase}$ denotes the proposed adaptive phase consistency loss for structural constraint (Eq.~\ref{eq:sc_loss}). Here, $(x_i^{l},y_i^{l})$ represents a labeled degraded image and its corresponding ground truth image, while $\alpha$ and $\beta$ are balancing coefficients. In our paper, both are empirically set to 0.2.

\subsection{Contrastive Learning for Unsupervised Learning}
Ideally, the teacher network produces higher-quality predictions than the student network and therefore provides reliable supervisory signals for unlabeled data. To improve the quality of pseudo-GTs, other methods \citep{Huang2023ContrastiveSL,Xu2024TowardsRA} retain only reliable teacher outputs, where reliability is assessed using the no-reference image quality assessment (NR-IQA) metric. However, this assumption does not always hold in practice. As shown in Fig.~\ref{fig:test_with_teacher}, the teacher may still generate unsatisfactory predictions containing severe artifacts, color distortion, or only marginal improvements over the degraded input. Such erroneous outputs are simply discarded in existing methods, leaving their valuable negative information unexplored.

To exploit these failure cases, we introduce an \emph{unreliable database} that explicitly stores the best historical failure for each unlabeled sample. Instead of treating poor predictions as useless, we regard them as informative negative samples that indicate undesirable restoration outcomes. During training, the student network is encouraged to avoid producing outputs similar to those stored in the unreliable database, while the teacher network inherits this behavior through EMA updates.

Specifically, we initialize both the reliable database and the unreliable database as empty. For each unlabeled image $x_i^u \in D_U=\{x_i^u\}_{i=1}^{M}$, the current teacher prediction is compared with the degraded input, the student prediction, and the historical entries in both databases. If the teacher prediction achieves the highest image quality, it replaces the corresponding entry in the reliable database. Otherwise, if it represents a failed prediction but is of higher quality than the existing entry in the unreliable database, the unreliable database is updated accordingly. The complete updating procedure is summarized in Algo.~\ref{alg:bankupdate}. Consequently, the reliable database $B_R=\{\hat{y}_i^r\}_{i=1}^{M}$ maintains the best pseudo-GT for each unlabeled image, whereas the unreliable database $B_U=\{\hat{y}_i^u\}_{i=1}^{M}$ records the strongest failure encountered during training, providing explicit negative supervision that discourages the model from repeating similar restoration errors.

With the reliable pseudo-GTs and unreliable samples available, the unsupervised appearance loss is formulated as follows:
\begin{equation}
    L_\text{app} = \sum_{i=1}^{M}|y_i^s-\hat{y}_i^r|,y_i^s= f_{\theta_s}(x_i^u) \>,
    \label{eq:app_loss}
\end{equation}
where $y_i^s$ is the restored student network output. To keep the restored output close to the reliable pseudo-GT and separated from the degraded input image and unreliable sample, we implement contrastive loss in the training
\begin{equation}
    L_\text{con}= \sum_{j=i}^{K}\sum_{i=1}^{M}\gamma_j\frac{e^{|\phi_j(y_i^s)-\phi_j(\hat{y}_i^r)|}}{\sum_{y\in \{\hat{y}_i^r,\hat{y}_i^u,x_i^u\}} e^{|\phi_j(y_i^s)-\phi_j(y)|}} \>,
    \label{eq:contrast_loss}
\end{equation}
where $\phi_{j}(\cdot)$ represents the $j_{th}$ hidden layer of the pre-trained VGG-19 \citep{Simonyan2014VeryDC}, and $\gamma_j$ is the weight coefficient. From Eq.\ \ref{eq:app_loss} and \ref{eq:contrast_loss}, we derive our loss function for unsupervised learning,
\begin{equation}
    L_\text{ul} = L_\text{con} + L_\text{app}
    \label{eq:unsup_loss}
\end{equation}

\begin{algorithm}[t]
\caption{Update of Reliable and Unreliable Database}\label{alg:bankupdate}
\begin{algorithmic}
\Require NR-IQA method $\text{sco}(\cdot)$, unlabeled  dataset $D_U$, batch size $b$, teacher network $f_{\theta_{t}}$, student network $f_{\theta_{s}}$, unreliable database $B_U$, and reliable database $B_R$ 
\Ensure
\State Initialize $B_U=\emptyset,B_{R}=\emptyset$
\For{batch $B=\{x_i^u\}^b_{i=1}$ in $D_U$}
    \For{$x_i^u$ in $B$}
    \State $y_{i}^{t} = f_{\theta_t}(x_i^u)$ \emph{\# teacher's output}
    \State $y_{i}^{s} = f_{\theta_s}(x_i^u)$ \emph{\# student's output}
    \State \emph{\# Compute NR-IQA scores of $x_i^u$, $y_{i}^{t}$, $y_{i}^{s}$, $\hat{y}_i^r\in B_R$, and $\hat{y}_i^u\in B_U$}
    \State $z^x_i = \text{sco}(x_i^u),z^t_i = \text{sco}(y_{i}^{t}),z^s_i = \text{sco}(y_{i}^{s})$
    \State $z^r_i = \text{sco}(\hat{y}_i^r),z^u_i = \text{sco}(\hat{y}_i^u)$
    \If{$z^t_i>z^x_i$ and $z^t_i>z^s_i$ and $z^t_i>z^r_i$}
        \State Replace $\hat{y}_i^r$ in $B_R$ by $y_{i}^{t}$
            \ElsIf{ $z^t_i>z^u_i$}
        \State Replace $\hat{y}_i^u$ in $B_U$ by $y_{i}^{t}$
    \EndIf
    \EndFor
\EndFor
\end{algorithmic}
\end{algorithm}

\subsection{Structural Constraint for Unsupervised Learning}
Restoring images degraded by adverse weather requires not only recovering visual fidelity but also preserving the underlying structures obscured by weather-induced artifacts. Semantic supervision is particularly beneficial in semi-supervised learning, where reliable labels are unavailable for real-world data. To exploit semantic information, WResVLM \citep{Xu2024TowardsRA} employs VLMs to construct a contrastive semantic supervision objective. Although effective, such approaches incur substantial computational overhead and rely on language representations that are inherently ambiguous and weakly aligned with the fine-grained spatial structures required for low-level vision tasks.

Instead, we propose a lightweight phase spectrum-based structural constraint. Our design is motivated by two observations. First, previous studies have shown that the Fourier phase spectrum preserves most of the semantic and structural information of an image \citep{Guo2022ExploringFP}. Second, although the phase spectrum is generally robust to illumination changes and color shifts, severe degradations may still corrupt its structural information \citep{Liu2025FrequencyDD}. Consequently, neither the degraded input nor the teacher-generated pseudo-GT should be treated as an absolutely reliable structural reference.

To address this issue, we construct an adaptive phase consistency loss by combining the phase spectrum of the degraded input and the reliable pseudo-GT according to their estimated image qualities. Specifically, the student prediction is encouraged to align its phase spectrum with this adaptive reference through the following phase consistency loss
\begin{equation}
    L_\text{phase} = \sum_{i=1}^{M} w_{i}d_\text{og} + (1-w_i)d_\text{oi} \>,
    \label{eq:sc_loss}
\end{equation}
where $P(\cdot)$ denotes the Fourier phase spectrum, $d_\text{og}=|P(y^{s}_i) - P(\hat{y}^r_i)|$ and $d_\text{oi}=|P(y^{s}_i) - P(x^u_i)|$ are L1 distance.
\begin{equation}
    w_i = \frac{z^r_i}{z^x_i + z^r_i} \>,
\end{equation}
where $z^r_i = \text{sco}(\hat{y}_i^r)$ and $z^x_i = \text{sco}(x_i^u)$ denote the NR-IQA scores of the reliable pseudo-GT and degraded input, respectively. Intuitively, severe degradations result in a significantly lower quality score for the input image than for the pseudo-GT, yielding a larger $w_i$
. In this case, the structural constraint relies more heavily on the reliable pseudo-GT. Conversely, when the degradation is mild, the degraded input already retains sufficient structural information, causing $w_i$ to decrease and the supervision to place greater emphasis on the input image. This adaptive strategy enables the model to exploit the most reliable structural information available under varying degradation levels while avoiding excessive dependence on either source alone.

\begin{table}[t]
\centering
\resizebox{\linewidth}{!}{
\renewcommand{\arraystretch}{1.15}
\begin{tabular}{ll|cc|cc|cc|cc|cc}
\Xhline{1.05pt}
\multicolumn{2}{c|}{Models}                          & \multicolumn{2}{c|}{Rain}       & \multicolumn{2}{c|}{Snow}       & \multicolumn{2}{c|}{Haze}       & \multicolumn{2}{c|}{Average}        &  \#Param & FLOP   \\
\multicolumn{1}{l}{Name} & \multicolumn{1}{l|}{Year} & LIQE$\uparrow$          & CLIP-IQA$\uparrow$      & LIQE$\uparrow$           & CLIP-IQA$\uparrow$      & LIQE$\uparrow$           & CLIP-IQA$\uparrow$       & LIQE$\uparrow$           & CLIP-IQA$\uparrow$      & M        & G      \\ \hline
Restormer                & CVPR22                    & 2.277          & 0.437          & 3.172          & 0.510          & 1.918          & 0.366          & 2.456          & 0.438          & 26.12    & 141.00 \\
TransWeather             & CVPR22                    & 1.924          & 0.358          & 2.770          & 0.416          & 1.502          & 0.292          & 2.065          & 0.355          & 38.05    & 12.24  \\
TKL                      & CVPR22                    & 2.028          & 0.392          & 2.830          & 0.428          & 1.590          & 0.318          & 2.149          & 0.379          & 28.47    & 49.22  \\
WeatherDiffusion              & TPAMI23                   & 2.050          & 0.395          & 2.950          & 0.466          & 1.520          & 0.326          & 2.173          & 0.396          & 82.96    & 475.16 \\
WGWS-Net                 & CVPR23                    & 1.965          & 0.389          & 2.619          & 0.395          & 1.506          & 0.310          & 2.030          & 0.365          & 5.19     & 13.61  \\
MWDT                     & ICCV23                    & 2.068          & 0.377          & 2.903          & 0.407          & 1.720          & 0.315          & 2.230          & 0.366          & -        & -      \\
PromptIR                 & NeurIPS23                 & 2.250          & 0.439          & 3.121          & 0.517          & 1.941          & 0.372          & 2.437          & 0.443          & 35.59    & 15.81  \\
DA-CLIP                  & ICLR24                    & 2.250          & 0.412          & 3.050          & 0.499          & 2.014          & 0.325          & 2.438          & 0.412          & 174.10   & 118.50 \\
WResVLM                  & ECCV24                    & 2.563          & 0.477          & 3.293          & 0.519          & 2.064          & 0.371          & 2.640          & 0.456          & -        & -      \\
Histoformer & ECCV24 & 2.563 & 0.443 & 3.273 & 0.513 & 1.982 & 0.386 & 2.606 & 0.447 & 16.62 & 91.59 \\
MODEM                    & NeurIPS25                 & \textbf{3.499} & 0.582          & 3.823          & 0.584          & \underline{2.832}    & 0.447          & \underline{3.385}    & 0.538          & 19.96    & 463.86 \\
HOGformer                & AAAI26                    & 3.396          & \underline{0.585}    & \underline{3.827}    & \textbf{0.612} & 2.659          & \underline{0.456}    & 3.294          & \underline{0.551}    & 16.65    & 91.77  \\ \hline
\textbf{Ours}            & \multicolumn{1}{l|}{-}    & \underline{3.466}    & \textbf{0.597} & \textbf{3.852} & \underline{0.611}    & \textbf{3.263} & \textbf{0.507} & \textbf{3.527} & \textbf{0.572} & 19.67    & 153.94 \\\Xhline{1.25pt}
\end{tabular}}
\caption{Quantitative comparison in all-in-one setting. \textbf{Bold} and \underline{Underlining} indicate the best and second best results, respectively.}
\label{tab:quantitative_results}
\end{table}
\begin{table}[t]

\footnotesize
\small 
\resizebox{0.3\linewidth}{!}{
\renewcommand{\arraystretch}{1.15}
\begin{tabular}{l|cc}
\hline
\multicolumn{1}{c|}{\multirow{2}{*}{Models}} & \multicolumn{2}{c}{RAIN100L}    \\
\multicolumn{1}{c|}{}                        & PSNR$\uparrow$           & SSIM$\uparrow$         \\ \hline
UMR                                          & 32.39          & 0.921          \\
SIRR                                         & 32.37          & 0.926          \\
MSPFN                                        & 33.50          & 0.948          \\
LPNet                                        & 33.61          & 0.958          \\
AirNet                                       & 34.90          & 0.977          \\
Restormer                                    & 36.74          & 0.978          \\
PromptIR                                     & 37.04          & 0.979          \\
DFPIR                                        & \textbf{39.08} & \underline{0.984}    \\ \hline
Ours                                         & \underline{39.03}    & \textbf{0.989}
\end{tabular}}
\hfill
\resizebox{0.35\linewidth}{!}{
\renewcommand{\arraystretch}{1.15}
\begin{tabular}{l|cc|cc}
\hline
\multicolumn{1}{c|}{\multirow{2}{*}{Models}} & \multicolumn{2}{c|}{Snow100K-S} & \multicolumn{2}{c}{Snow100K-L}  \\
\multicolumn{1}{c|}{}                        & PSNR$\uparrow$          & SSIM$\uparrow$           & PSNR$\uparrow$          & SSIM$\uparrow$          \\ \hline
MPRNet                                       & 36.22          & 0.973          & 30.63          & 0.938          \\
SwinIR                                       & 33.45          & 0.960          & 28.40          & 0.915          \\
Restormer                                    & 36.50          & 0.973          & 30.86          & 0.938          \\
NAFNet                                       & 35.94          & 9.970          & 29.50          & 0.924          \\
UniRestorer                                  & \underline{36.94}    & \underline{0.974}    & \underline{31.14}    & \underline{0.940}    \\ \hline
Ours                                         & \textbf{37.07} & \textbf{0.981} & \textbf{31.38} & \textbf{0.949}
\end{tabular}}
\hfill
\resizebox{0.3\linewidth}{!}{
\renewcommand{\arraystretch}{1.15}
\begin{tabular}{l|cc}
\hline
\multirow{2}{*}{Models} & \multicolumn{2}{c}{SOTS}        \\
                        & PSNR$\uparrow$          & SSIM$\uparrow$          \\ \hline
DehazeNet               & 22.46          & 0.851          \\
MSCNN                   & 22.06          & 0.908          \\
AODNet                  & 20.29          & 0.877          \\
EPDN                    & 22.57          & 0.863          \\
FDGAN                   & 23.15          & 0.921          \\
AirNet                  & 23.18          & 0.900          \\
Restormer               & 30.87          & 0.969          \\
DFPIR                   & \underline{32.00}    & \textbf{0.981} \\ \hline
Ours                    & \textbf{32.03} & \underline{0.980}
\end{tabular}}

\caption{Quantitative comparison in single-task setting. From left to right: deraining, desnowing, and dehazing.}
\label{tab:single_task}
\end{table}

\begin{figure*}[!t]
  \centering

    \includegraphics[width=\linewidth]{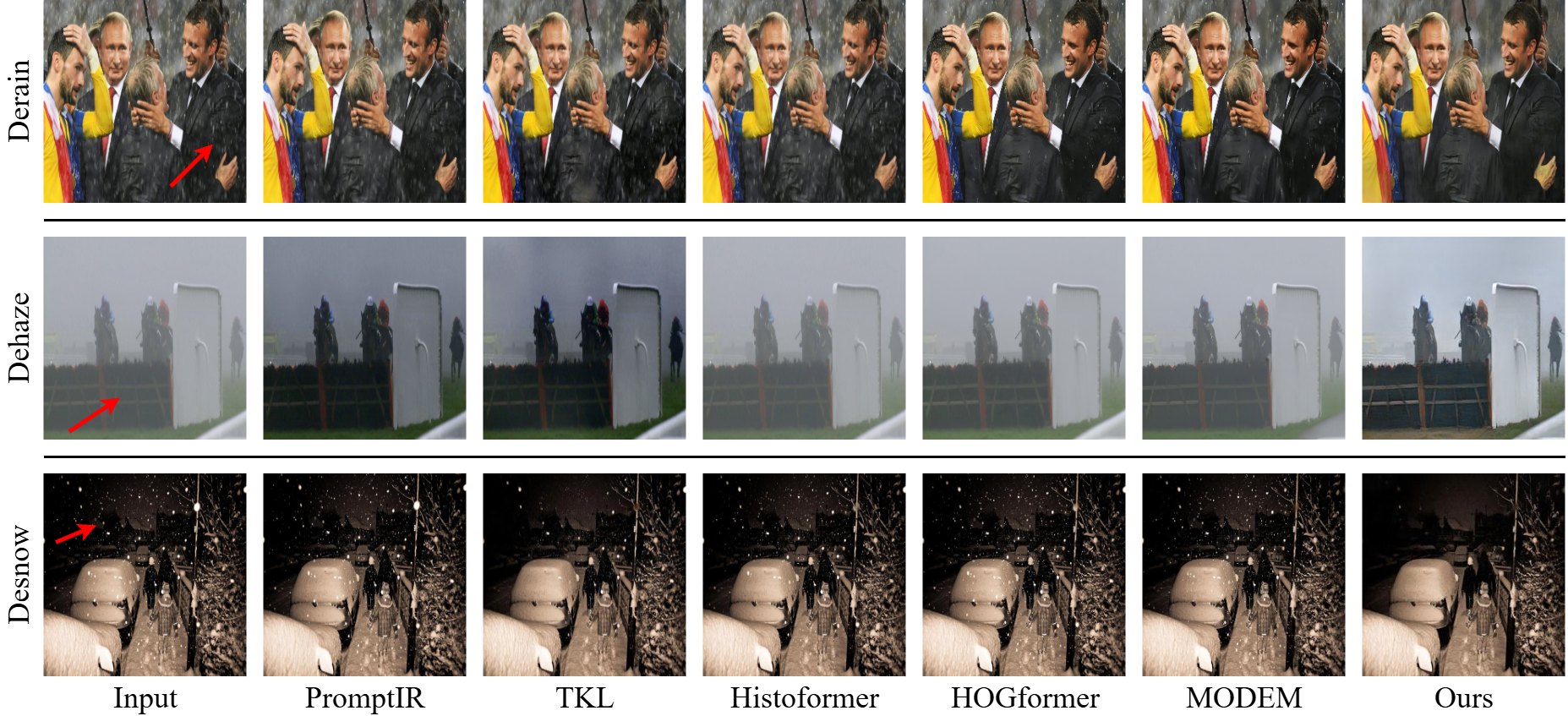}   
    \caption{Qualitative comparisons of adverse weather restoration. The \textcolor{red}{red} arrows highlight regions where the proposed method produces superior visual results. Please zoom in for better visualization.}

  \label{fig:qualitative_results}
\end{figure*}

\section{Experimental Results}
In this section, we present the experimental results of our method. We train our method in two settings: all-in-one and single-task, with the same setup. Our semi-supervised learning framework is easily compatible with various image restoration networks. Following \citep{Xu2024TowardsRA}, we opt for MSBDN \citep{Dong2020MultiScaleBD} as our backbone due to its balanced performance and rapid inference speed in our main study. For the NR-IQA method, we incorporate the SOTA RALI score \citep{zhao2025reasoning}. Each batch comprises $8$ labeled and $8$ unlabeled images, and the number of epochs is set to $200$. During training, we used the Adam optimizer \citep{Kingma2014AdamAM} with $\beta_{1}=0.9$ and $\beta_{2}=0.999$ and the learning rate $2e-5$. All training was conducted on $2\times H100$.
\subsection{Datasets}
Following previous works \citep{Ye_2023_ICCV,Sun2024RestoringII,Wang2025MODEMAM}, we adopt the all-weather dataset \citep{Ye_2023_ICCV} as labeled dataset. Meanwhile, we leverage the unlabeled real-world adverse weather images for unsupervised learning. To achieve this, we utilize the real hazy image in URHI \citep{Li2017BenchmarkingSD}, the real rainy image in the SPA+ test set \citep{zhu2023Weather}, and the real snowy image from the RealSnow dataset \citep{zhu2023Weather}. Following \citep{Xu2024TowardsRA}, we employ real adverse weather image datasets for qualitative and quantitative evaluation for the unified model, including RTTS \citep{Li2017BenchmarkingSD}, DDN-SIRR \citep{Wei_2019_CVPR}, and Real3000 \citep{9711111}, and Snow100k Realistic \citep{Liu2017DesnowNetCD}. As ground truth is not available, we employ widely used non-reference metrics LIQE \citep{zhang2023liqe} and CLIP-IQA \citep{wang2022exploring}. 

For the model in the single-task setting, $i.e.$ separate models are trained for different restoration tasks, we use Rain100L \citep{Ye_2023_ICCV} for rain, Snow100K-S and Snow100K-L \citep{Liu2017DesnowNetCD} for snow, and SOTS \citep{Li2017BenchmarkingSD} for haze. The results are reported in PSNR and SSIM.

\subsection{Quantitative Results}
The results of competing methods are obtained either by directly citing the reported results from the original papers, reproducing the results using the authors' publicly released models, or retraining the models using their official implementations.

We compare the proposed method with recent state-of-the-art all-in-one image restoration models, including Restormer \citep{Zamir2021RestormerET}, TransWeather \citep{Ye_2023_ICCV}, TKL \citep{Chen2022LearningMA}, WeatherDiffusion \citep{Sun2024RestoringII}, WGWS-Net \citep{zhu2023Weather}, MWDT \citep{10377637}, PromptIR \citep{Potlapalli2023PromptIRPF}, DA-CLIP \citep{Luo2023ControllingVM}, WResVLM \citep{Xu2024TowardsRA}, Histoformer \citep{Sun2024RestoringII}, MODEM \citep{Wang2025MODEMAM}, and HOGformer \citep{Wu2025GradientAC}. The quantitative comparisons are presented in Tab.~\ref{tab:quantitative_results}. Our method achieves the best overall performance, surpassing the second-best method by 0.142 in LIQE and 0.021 in CLIP-IQA. Moreover, within the family of semi-supervised methods, our approach consistently outperforms WResVLM across all evaluated weather restoration tasks. Despite these improvements, our method remains computationally efficient. Compared with MODEM, it achieves comparable model complexity with the same number of parameters while requiring only approximately one-third of the computational cost (GFLOPs).

To further evaluate the generalization ability of the proposed framework, we also compare it with representative single-task image restoration methods, including UMR \citep{Yasarla2019UncertaintyGM}, SIRR \citep{Wei_2019_CVPR}, MSPFN \citep{Jiang2020MultiScalePF}, DehazeNet \citep{Cai2016DehazeNetAE}, MSCNN \citep{Ren2016SingleID}, AODNet \citep{Li2017AODNetAD}, EPDN \citep{Qu2019EnhancedPD}, FDGAN \citep{Dong2020FDGANGA}, LPNet \citep{Gao2019DynamicSD}, AirNet \citep{Li2022AllInOneIR}, Restormer \citep{Zamir2021RestormerET}, PromptIR \citep{Potlapalli2023PromptIRPF}, SwinIR \citep{Liang2021SwinIRIR}, NAFNet \citep{Chen2022SimpleBF}, UniRestorer \citep{lin2026unirestorer}, and DFPIR \citep{tian2025degradation}. The results are summarized in Tab.~\ref{tab:single_task}. Our method achieves the highest PSNR on the desnowing task, outperforming UniRestorer by at least 0.13 dB. On the remaining restoration tasks, our method consistently delivers competitive performance, closely matching or exceeding the recent diffusion-based approach DFPIR. In this setting, the unsupervised branch of our framework is trained using the same synthetic dataset as the supervised branch with the GT stored in a reliable database.
More quantitative results are provided in Appendix~\ref{app:od_task} and \ref{app:diff_metrics}.
\subsection{Qualitative Results}
We qualitatively compare our method with state-of-the-art image restoration approaches on challenging real-world adverse weather images. The visual comparisons are presented in Fig.~\ref{fig:qualitative_results}. As can be observed, most competing methods struggle to simultaneously remove rain, haze, and snow artifacts, often leaving noticeable degradations or introducing undesirable visual distortions. In particular, PromptIR \citep{Potlapalli2023PromptIRPF} and TKL \citep{Chen2022LearningMA} partially suppress haze but tend to produce texture distortions and over-smoothed details. In contrast, our method consistently restores clearer scene structures while preserving natural textures, resulting in superior visual quality with enhanced clarity, contrast, and overall perceptual realism. Moreover, our approach effectively removes residual haze commonly present in rainy and snowy scenes, leading to significantly improved visibility.

\subsection{Quantitative Results with Composited Dataset}
\begin{table}[t]
\centering
\resizebox{\linewidth}{!}{
\renewcommand{\arraystretch}{1.15}
\begin{tabular}{l|ccccccc}
\hline
\multirow{2}{*}{Models} & H                     & S                     & R                     & H+S                   & H+R                   & R*+S                  & H+S+R*                \\
                        & PSNR$\uparrow$/SSIM$\uparrow$             & PSNR$\uparrow$/SSIM$\uparrow$             & PSNR$\uparrow$/SSIM$\uparrow$             & PSNR$\uparrow$/SSIM$\uparrow$             & PSNR$\uparrow$/SSIM$\uparrow$             & PSNR$\uparrow$/SSIM$\uparrow$             & PSNR$\uparrow$/SSIM$\uparrow$             \\ \hline
MODEM                   & \underline{19.73}/\underline{0.8789}    & 27.72/0.8953          & \underline{23.76}/\underline{0.9029}    & \textbf{17.87}/\textbf{0.7918} & 20.16/0.8801          & 20.34/0.6734          & \underline{14.23}/0.4327 \\
HOGFormer               & 18.87/0.8693          & \underline{28.96}/\underline{0.9129}    & 23.54/0.9051          & 16.51/0.7635          & \underline{20.30}/\underline{0.8876}    & \underline{20.41}/\underline{0.6771}    & 13.66/\underline{0.4557}          \\ \hline
Ours                    & \textbf{19.89}/\textbf{0.8805} & \textbf{33.46}/\textbf{0.9568} & \textbf{24.01}/\textbf{0.9336} & \underline{17.27}/\underline{0.7874}    & \textbf{21.36}/\textbf{0.9161} & \textbf{22.84}/\textbf{0.7390} & \textbf{14.45}/\textbf{0.5012}   
\end{tabular}}
\caption{Result with composited datasets. * indicates that this degradation is synthetically added by us.}
\label{tab:composited_data}
\end{table}

Following \citep{Wu2025GradientAC}, we also conduct our experiments with the composited CDD11 dataset \citep{guo2024onerestore}, which includes several types of degradation and their combination. As the CDD11 dataset lacks some combinations of degradation that we are interested in testing, we have synthetically added them into existing combinations. 

\begin{figure}[b]
        \centering
        \includegraphics[width=0.6\linewidth]{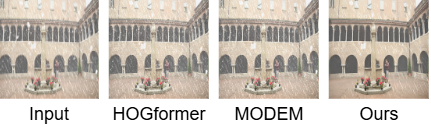}
        \caption{Comparisons on multiple degradations (rain-haze-snow).}
        \label{fig:test_with_composite}
    \end{figure}

In this experiment, we compare our method with HOGformer and MODEM unified models. We have reported the test in Tab.\ \ref{tab:composited_data} and Fig.\ \ref{fig:test_with_composite}, showing that our method consistently outperforms the two models in many combinations of degradation. Note that all models are not trained on the CDD11 dataset.

\subsection{Ablation Studies}
\noindent\textbf{Impact of components} 
To evaluate the effectiveness of the proposed components, we conduct ablation studies on three benchmark restoration tasks. In these experiments, the baseline model is the semi-supervised framework without $L_{con}$ (Eq.\ \ref{eq:contrast_loss}) and phase consistency loss (Eq.\ \ref{eq:sc_loss}). The average results of three tasks (desnowing, dehazing, and deraining) are reported in Tab.~\ref{tab:ab_study}. As shown, the proposed contrastive loss consistently improves the baseline performance. Introducing the unreliable database further enhances restoration quality, with an improvement of roughly $0.7$ in LIQE and $0.08$ in CLIP-IQA. The proposed phase spectrum-based constraint yields an additional improvement compared with the text-based (aligning the textual embedding of the restored output and reliable pseudo-GT with LLaVA \citep{Xu2024TowardsRA}) and pixel-based (LPIPS of restored output and reliable pseudo-GT) structural constraints. 
\begin{table}[t]
\centering
\resizebox{\linewidth}{!}{
\renewcommand{\arraystretch}{1.15}
\begin{tabular}{c|lc|lll|cc}
\hline
\multirow{2}{*}{Baseline} & \multicolumn{2}{c|}{\textit{w/} $L_{con}$}                                                                                                                 & \multicolumn{3}{c|}{\textit{w/} unsupervised constraint}                   & \multirow{2}{*}{LIQE $\uparrow$} & \multirow{2}{*}{CLIP-IQA $\uparrow$} \\ \cline{2-6}
                          & \multicolumn{1}{c}{\begin{tabular}[c]{@{}c@{}}\textit{w/o} unreliable\\ database\end{tabular}} & \begin{tabular}[c]{@{}c@{}}\textit{w/} unreliable\\ database\end{tabular} & \multicolumn{1}{c}{LPIPS} & \multicolumn{1}{c}{Text} & \multicolumn{1}{c|}{Phase} &                       &                           \\ \hline
x                         &                                                                                       & \multicolumn{1}{l|}{}                                            &                           &                          &       & 2.390                 & 0.402                     \\
x                         & \multicolumn{1}{c}{x}                                                                 & \multicolumn{1}{c|}{}                                            &                           &                          &       & 2.429                 & 0.445                     \\
x                         &                                                                                       & x                                                                &                           &                          &       & 3.098                 & 0.525                     \\
x                         &                                                                                       & x                                                                & \multicolumn{1}{c}{x}     &                          &       &     3.368                  &             0.565              \\
x                         &                                                                                       & x                                                                &                           & \multicolumn{1}{c}{x}    &       & 3.407                 & 0.545                     \\
x                         &                                                                                       & x                                                                &                           &                          & \multicolumn{1}{c|}{x}     & \textbf{3.527}        & \textbf{0.572}           
\end{tabular}
}
\caption{Ablation study on effectiveness of proposed components. We report the average results of three tasks (desnowing, dehazing, and deraining).}
\label{tab:ab_study}
\end{table}

\noindent\textbf{Choice of NR-IQA Metric} Since the proposed framework relies on NR-IQA to assess pseudo-GT quality, the choice of metric is crucial. We compare several NR-IQA methods, and the results are summarized in Tab.~\ref{tab:nr_iqa}. Among all candidates, RELI consistently achieves the best performance across all restoration tasks, justifying its use in our framework.
\begin{table}[t]
\centering
\renewcommand{\arraystretch}{1.15}
\begin{tabular}{l|cc|cc|cc}
\hline
\multirow{2}{*}{NR-IQA} & \multicolumn{2}{c|}{Rain}        & \multicolumn{2}{c|}{Snow}       & \multicolumn{2}{c}{Haze}        \\
                        & LIQE$\uparrow$          & CLIP-IQA$\uparrow$       & LIQE$\uparrow$          & CLIP-IQA$\uparrow$       & LIQE$\uparrow$          & CLIP-IQA$\uparrow$      \\ \hline
LIQE                   & 1.029          & 0.418          & 1.013          & 0.389          & 1.032          & 0.453          \\
MUSIQ                   & 1.774          & 0.469          & 2.507          & 0.532          & 1.329          & 0.420          \\
NIMA                    & 2.194          & 0.516          & 2.228          & \underline{0.610}          & \underline{2.995}          & 0.468          \\
CLIP-IQA                & \underline{2.445}          & \underline{0.553}          & \underline{3.120}          & 0.603          & 1.696          & \textbf{0.508}         \\
RALI                    & \textbf{3.466} & \textbf{0.597} & \textbf{3.852} & \textbf{0.611} & \textbf{3.263} & \underline{0.507}
\end{tabular}
\caption{Performance comparison of different NR-IQA methods.}
\label{tab:nr_iqa}
\end{table}
\begin{table}[!h]
\centering
\renewcommand{\arraystretch}{1.15}
\begin{tabular}{c|c|ccc}
\hline
\multirow{2}{*}{\begin{tabular}[c]{@{}c@{}}Degradation\\ types\end{tabular}} & \multirow{2}{*}{Metrics} & \multicolumn{3}{c}{Choices of $L_{sc}$}                         \\
                                   &                          & L1     & Contrastive & Ours            \\ \hline
\multirow{2}{*}{Rain}              & LIQE$\uparrow$                     & 3.302       & \underline{3.380}      & \textbf{3.466} \\
                                   & CLIP-IQA$\uparrow$                 & 0.592       & \underline{0.593}      & \textbf{0.597} \\ \hline
\multirow{2}{*}{Snow}              & LIQE$\uparrow$                     & 3.742       & \textbf{3.862}   & \underline{3.852}    \\
                                   & CLIP-IQA$\uparrow$                 & 0.596       & \textbf{0.612}   & \underline{0.611}    \\ \hline
\multirow{2}{*}{Haze}              & LIQUE$\uparrow$                   & 3.136       & \underline{3.192}      & \textbf{3.263} \\
                                   & CLIP-IQA$\uparrow$                & \underline{0.504} & 0.489            & \textbf{0.507} \\ \hline
\multirow{2}{*}{Average}               & LIQE$\uparrow$                    & 3.394       & \underline{3.478}      & \textbf{3.527} \\
                                   & CLIP-IQA$\uparrow$                & 0.564       & \underline{0.565}      & \textbf{0.572}
\end{tabular}
\caption{Performance comparison with different $L_\text{phase}$.}
\label{tab:l_sc}
\end{table}

\noindent\textbf{Effectiveness of Adaptive Phase consistency loss} To validate the proposed adaptive phase consistency loss, we compare it with alternative unsupervised constraints, including an $L_1$ loss using the reliable pseudo-GT as the supervision target and a contrastive loss using the reliable pseudo-GT and degraded input as the positive and negative samples, respectively. The results are summarized in Tab.~\ref{tab:l_sc}. As shown, even though the contrastive loss performs better in the desnowing task, the proposed adaptive phase consistency loss consistently achieves the best overall performance.


Further ablation studies presented in the Appendix include phase visualization (\ref{app:spectral}), overhead comparison with other SOTAs (\ref{app:overhead_sota}), overhead comparison with other constraint methods (\ref{app:overhead_unsupervised_constraints}), ablation studies on backbone models (\ref{acc:backbones}) and adaptive weight (\ref{app:adaptive_weight}), and impact of unreliable data (\ref{app:unreliable}).

\section{Conclusion}
This paper presented an unreliable-aware semi-supervised learning framework for real-world adverse weather image restoration. The proposed framework combines an unreliable bank for exploiting failed pseudo-GTs as hard negative samples with an adaptive phase consistency loss that replaces expensive vision-language model-based semantic supervision. Together, these components effectively reduce confirmation bias while providing efficient structural guidance during training. Future work will investigate extending the implementation.



\subsection*{AI use statement}

In this work, we used generative AI tools for cleaning and reformatting the dataset, grammar checking. We have not used generative AI tools to generate synthetic data sets, help develop theoretical models or conceptual frameworks, formulate mathematical claims, provide critical ingredients for proving mathematical claims, propose or refine hypotheses, design or provide feedback on research methodology or experiments, implement methods, assist with translation, support qualitative and interpret results, assist in the writing of proofs and polish the manuscript. Thematic data analysis are not applicable to this work
Additionally, we used generative AI tools to propose a title or keywords for a research paper. We have reviewed all AI-assisted work. We take responsibility for the final content of this work,
including text, claims or artifacts produced with the aid of generative AI.

\bibliography{iclr2027_conference}

@article{Yasarla2019UncertaintyGM,
  title={Uncertainty Guided Multi-Scale Residual Learning-Using a Cycle Spinning CNN for Single Image De-Raining},
  author={Rajeev Yasarla and Vishal M. Patel},
  journal={2019 IEEE/CVF Conference on Computer Vision and Pattern Recognition (CVPR)},
  year={2019},
  pages={8397-8406},
  url={https://api.semanticscholar.org/CorpusID:195657934}
}

@misc{tang2024understandingperformancegaponline,
      title={Understanding the performance gap between online and offline alignment algorithms}, 
      author={Yunhao Tang and Daniel Zhaohan Guo and Zeyu Zheng and Daniele Calandriello and Yuan Cao and Eugene Tarassov and Rémi Munos and Bernardo Ávila Pires and Michal Valko and Yong Cheng and Will Dabney},
      year={2024},
      eprint={2405.08448},
      archivePrefix={arXiv},
      primaryClass={cs.LG},
      url={https://arxiv.org/abs/2405.08448}, 
}

@inproceedings{guo2024onerestore,
  title={OneRestore: A Universal Restoration Framework for Composite Degradation},
  author={Guo, Yu and Gao, Yuan and Lu, Yuxu and Liu, Ryan Wen and He, Shengfeng},
  booktitle={European Conference on Computer Vision},
  year={2024}
}

@article{wu2023qalign,
  title={Q-Align: Teaching LMMs for Visual Scoring via Discrete Text-Defined Levels},
  author={Wu, Haoning and Zhang, Zicheng and Zhang, Weixia and Chen, Chaofeng and Li, Chunyi and Liao, Liang and Wang, Annan and Zhang, Erli and Sun, Wenxiu and Yan, Qiong and Min, Xiongkuo and Zhai, Guangtai and Lin, Weisi},
  journal={arXiv preprint arXiv:2312.17090},
  year={2023},
  institution={Nanyang Technological University and Shanghai Jiao Tong University and Sensetime Research},
  note={Equal Contribution by Wu, Haoning and Zhang, Zicheng. Project Lead by Wu, Haoning. Corresponding Authors: Zhai, Guangtai and Lin, Weisi.}
}

@inproceedings{
lin2026unirestorer,
title={UniRestorer: Universal Image Restoration via Adaptively Estimating Image Degradation at Proper Granularity},
author={Jingbo Lin and Zhilu Zhang and Wenbo Li and Renjing Pei and Hang Xu and Hongzhi Zhang and Wangmeng Zuo},
booktitle={The Fourteenth International Conference on Learning Representations},
year={2026},
url={https://openreview.net/forum?id=nDrZow7fCF}
}

@article{zhao2025reasoning,
  title={Reasoning as Representation: Rethinking Visual Reinforcement Learning in Image Quality Assessment},
  author={Zhao, Shijie and Zhang, Xuanyu and Li, Weiqi and Li, Junlin and Zhang, Li and Xue, Tianfan and Zhang, Jian},
  journal={Proceedings of the International Conference on Learning Representations (ICLR)},
  year={2026}
}

@inproceedings{Chen2022SimpleBF,
  title={Simple Baselines for Image Restoration},
  author={Liangyu Chen and Xiaojie Chu and X. Zhang and Jian Sun},
  booktitle={European Conference on Computer Vision},
  year={2022},
  url={https://api.semanticscholar.org/CorpusID:248085491}
}

@inproceedings{tian2025degradation,
  title={Degradation-Aware Feature Perturbation for All-in-One Image Restoration},
  author={Tian, Xiangpeng and Liao, Xiangyu and Liu, Xiao and Li, Meng and Ren, Chao},
  booktitle={Proceedings of the Computer Vision and Pattern Recognition Conference},
  pages={28165--28175},
  year={2025}
}

@article{Li2017AODNetAD,
  title={AOD-Net: All-in-One Dehazing Network},
  author={Boyi Li and Xiulian Peng and Zhangyang Wang and Jizheng Xu and Dan Feng},
  journal={2017 IEEE International Conference on Computer Vision (ICCV)},
  year={2017},
  pages={4780-4788},
  url={https://api.semanticscholar.org/CorpusID:30151664}
}

@article{Cai2016DehazeNetAE,
  title={DehazeNet: An End-to-End System for Single Image Haze Removal},
  author={Bolun Cai and Xiangmin Xu and Kui Jia and Chunmei Qing and Dacheng Tao},
  journal={IEEE Transactions on Image Processing},
  year={2016},
  volume={25},
  pages={5187-5198},
  url={https://api.semanticscholar.org/CorpusID:14092238}
}

@article{Qu2019EnhancedPD,
  title={Enhanced Pix2pix Dehazing Network},
  author={Yanyun Qu and Yizi Chen and Jingying Huang and Yuan Xie},
  journal={2019 IEEE/CVF Conference on Computer Vision and Pattern Recognition (CVPR)},
  year={2019},
  pages={8152-8160},
  url={https://api.semanticscholar.org/CorpusID:195489762}
}

@article{Dong2020FDGANGA,
  title={FD-GAN: Generative Adversarial Networks with Fusion-discriminator for Single Image Dehazing},
  author={Yu Dong and Yihao Liu and He Zhang and Shifeng Chen and Yu Qiao},
  journal={ArXiv},
  year={2020},
  volume={abs/2001.06968},
  url={https://api.semanticscholar.org/CorpusID:210838848}
}

@inproceedings{Ren2016SingleID,
  title={Single Image Dehazing via Multi-scale Convolutional Neural Networks},
  author={Wenqi Ren and Sibo Liu and Hua Zhang and Jin-shan Pan and Xiaochun Cao and Ming-Hsuan Yang},
  booktitle={European Conference on Computer Vision},
  year={2016},
  url={https://api.semanticscholar.org/CorpusID:17763780}
}

@article{Jiang2020MultiScalePF,
  title={Multi-Scale Progressive Fusion Network for Single Image Deraining},
  author={Kui Jiang and Zhongyuan Wang and Peng Yi and Chen Chen and Baojin Huang and Yimin Luo and Jiayi Ma and Junjun Jiang},
  journal={2020 IEEE/CVF Conference on Computer Vision and Pattern Recognition (CVPR)},
  year={2020},
  pages={8343-8352},
  url={https://api.semanticscholar.org/CorpusID:214623217}
}

@article{Gao2019DynamicSD,
  title={Dynamic Scene Deblurring With Parameter Selective Sharing and Nested Skip Connections},
  author={Hongyun Gao and Xin Tao and Xiaoyong Shen and Jiaya Jia},
  journal={2019 IEEE/CVF Conference on Computer Vision and Pattern Recognition (CVPR)},
  year={2019},
  pages={3843-3851},
  url={https://api.semanticscholar.org/CorpusID:198185751}
}

@article{Li2022AllInOneIR,
  title={All-In-One Image Restoration for Unknown Corruption},
  author={Boyun Li and Xiao Liu and Peng Hu and Zhongqin Wu and Jiancheng Lv and Xiaocui Peng},
  journal={2022 IEEE/CVF Conference on Computer Vision and Pattern Recognition (CVPR)},
  year={2022},
  pages={17431-17441},
  url={https://api.semanticscholar.org/CorpusID:250551851}
}

@inproceedings{zhang2023liqe,  
  title={Blind Image Quality Assessment via Vision-Language Correspondence: A Multitask Learning Perspective},  
  author={Zhang, Weixia and Zhai, Guangtao and Wei, Ying and Yang, Xiaokang and Ma, Kede},  
  booktitle={IEEE Conference on Computer Vision and Pattern Recognition},  
  pages={14071--14081},
  year={2023}
}

@inproceedings{wang2022exploring,
    author = {Wang, Jianyi and Chan, Kelvin CK and Loy, Chen Change},
    title = {Exploring CLIP for Assessing the Look and Feel of Images},
    booktitle = {AAAI},
    year = {2023}
}

@article{Dong2020MultiScaleBD,
  title={Multi-Scale Boosted Dehazing Network With Dense Feature Fusion},
  author={Hang Dong and Jin-shan Pan and Lei Xiang and Zhe Hu and Xinyi Zhang and Fei Wang and Ming-Hsuan Yang},
  journal={2020 IEEE/CVF Conference on Computer Vision and Pattern Recognition (CVPR)},
  year={2020},
  pages={2154-2164},
  url={https://api.semanticscholar.org/CorpusID:216562731}
}

@article{Kingma2014AdamAM,
  title={Adam: A Method for Stochastic Optimization},
  author={Diederik P. Kingma and Jimmy Ba},
  journal={CoRR},
  year={2014},
  volume={abs/1412.6980},
  url={https://api.semanticscholar.org/CorpusID:6628106}
}

@INPROCEEDINGS{9711111,
  author={Liu, Yang and Yue, Ziyu and Pan, Jinshan and Su, Zhixun},
  booktitle={2021 IEEE/CVF International Conference on Computer Vision (ICCV)}, 
  title={Unpaired Learning for Deep Image Deraining with Rain Direction Regularizer}, 
  year={2021},
  volume={},
  number={},
  pages={4733-4741},
  doi={10.1109/ICCV48922.2021.00471}}

@article{Liu2017DesnowNetCD,
  title={DesnowNet: Context-Aware Deep Network for Snow Removal},
  author={Yun-Fu Liu and Da-Wei Jaw and Shih-Chia Huang and Jenq-Neng Hwang},
  journal={IEEE Transactions on Image Processing},
  year={2017},
  volume={27},
  pages={3064-3073},
  url={https://api.semanticscholar.org/CorpusID:837707}
}

@InProceedings{Wei_2019_CVPR,
author = {Wei, Wei and Meng, Deyu and Zhao, Qian and Xu, Zongben and Wu, Ying},
title = {Semi-Supervised Transfer Learning for Image Rain Removal},
booktitle = {The IEEE Conference on Computer Vision and Pattern Recognition},
year = {2019}
}

@article{Li2017BenchmarkingSD,
  title={Benchmarking Single-Image Dehazing and Beyond},
  author={Boyi Li and Wenqi Ren and Dengpan Fu and Dacheng Tao and Dan Feng and Wenjun Zeng and Zhangyang Wang},
  journal={IEEE Transactions on Image Processing},
  year={2017},
  volume={28},
  pages={492-505},
  url={https://api.semanticscholar.org/CorpusID:39760169}
}

@article{Simonyan2014VeryDC,
  title={Very Deep Convolutional Networks for Large-Scale Image Recognition},
  author={Karen Simonyan and Andrew Zisserman},
  journal={CoRR},
  year={2014},
  volume={abs/1409.1556},
  url={https://api.semanticscholar.org/CorpusID:14124313}
}

@article{Ke2021MUSIQMI,
  title={MUSIQ: Multi-scale Image Quality Transformer},
  author={Junjie Ke and Qifei Wang and Yilin Wang and Peyman Milanfar and Feng Yang},
  journal={2021 IEEE/CVF International Conference on Computer Vision (ICCV)},
  year={2021},
  pages={5128-5137},
  url={https://api.semanticscholar.org/CorpusID:237048383}
}

@article{Sohn2020FixMatchSS,
  title={FixMatch: Simplifying Semi-Supervised Learning with Consistency and Confidence},
  author={Kihyuk Sohn and David Berthelot and Chun-Liang Li and Zizhao Zhang and Nicholas Carlini and Ekin Dogus Cubuk and Alexey Kurakin and Han Zhang and Colin Raffel},
  journal={Conference on Neural Information Processing Systems},
  year={2020},
  volume={abs/2001.07685},
  url={https://api.semanticscholar.org/CorpusID:210839228}
}

@INPROCEEDINGS{11460995,
  author={Tang, Yunyi and Fang, Bowei and Zhao, Chunyu and Liu, Haoran and Yan, Fei and Deng, Tao},
  booktitle={ICASSP 2026 - 2026 IEEE International Conference on Acoustics, Speech and Signal Processing (ICASSP)}, 
  title={Adverse Effect Removal Network via Unsupervised Weather Type Transfer}, 
  year={2026},
  volume={},
  number={},
  pages={10057-10061},
  doi={10.1109/ICASSP55912.2026.11460995}}

@article{Liu2025RealWorldAW,
  title={Real-World Adverse Weather Image Restoration via Dual-Level Reinforcement Learning with High-Quality Cold Start},
  author={Fuyang Liu and Jiaqi Xu and Xiaowei Hu},
  journal={Conference on Neural Information Processing Systems},
  year={2025},
  volume={abs/2511.05095},
  url={https://api.semanticscholar.org/CorpusID:282890021}
}

@inproceedings{zhu2023Weather,
  title={Learning Weather-General and Weather-Specific Features for Image Restoration Under Multiple Adverse Weather Conditions},
  author={Yurui Zhu and Tianyu Wang and Xueyang Fu and Xuanyu Yang and Xin Guo and Jifeng Dai and Yu Qiao and Xiaowei Hu},
  booktitle={Proc. IEEE/CVF Conference on Computer Vision and Pattern Recognition (CVPR)},
  year={2023}
}

@article{Ye_2023_ICCV,
  title={TransWeather: Transformer-based Restoration of Images Degraded by Adverse Weather Conditions},
  author={Jeya Maria Jose Valanarasu and Rajeev Yasarla and Vishal M. Patel},
  journal={2022 IEEE/CVF Conference on Computer Vision and Pattern Recognition (CVPR)},
  year={2021},
  pages={2343-2353},
  url={https://api.semanticscholar.org/CorpusID:244714491}
}

@inproceedings{Sun2024RestoringII,
  title={Restoring Images in Adverse Weather Conditions via Histogram Transformer},
  author={Shangquan Sun and Wenqi Ren and Xinwei Gao and Rui Wang and Xiaochun Cao},
  booktitle={European Conference on Computer Vision},
  year={2024},
  url={https://api.semanticscholar.org/CorpusID:271212187}
}

@article{Zhang2022DesnowFormerAE,
  title={DesnowFormer: an effective transformer-based image desnowing network},
  author={Ting Zhang and Nanfeng Jiang and Junhong Lin and Jielian Lin and Tiesong Zhao},
  journal={2022 IEEE International Conference on Visual Communications and Image Processing (VCIP)},
  year={2022},
  pages={1-5},
  url={https://api.semanticscholar.org/CorpusID:255995928}
}

@article{song2023vision,
  title={Vision Transformers for Single Image Dehazing},
  author={Song, Yuda and He, Zhuqing and Qian, Hui and Du, Xin},
  journal={IEEE Transactions on Image Processing},
  year={2023},
  volume={32},
  pages={1927-1941}
}

@article{Chen2023LearningAS,
  title={Learning A Sparse Transformer Network for Effective Image Deraining},
  author={Xiang Chen and Hao‐Ran Li and Mingqiang Li and Jin-shan Pan},
  journal={2023 IEEE/CVF Conference on Computer Vision and Pattern Recognition (CVPR)},
  year={2023},
  pages={5896-5905},
  url={https://api.semanticscholar.org/CorpusID:257636509}
}

@article{Yang2024DepthAwareUV,
  title={Depth-Aware Unpaired Video Dehazing},
  author={Yang Yang and Chun-Le Guo and Xiaojie Guo},
  journal={IEEE Transactions on Image Processing},
  year={2024},
  volume={33},
  pages={2388-2403},
  url={https://api.semanticscholar.org/CorpusID:268635792}
}

@article{Yang2023RobustUI,
  title={Robust Unpaired Image Dehazing via Density and Depth Decomposition},
  author={Yang Yang and Chaoyue Wang and Xiaojie Guo and Dacheng Tao},
  journal={International Journal of Computer Vision},
  year={2023},
  volume={132},
  pages={1557 - 1577},
  url={https://api.semanticscholar.org/CorpusID:265452111}
}

@article{Liang2021SwinIRIR,
  title={SwinIR: Image Restoration Using Swin Transformer},
  author={Jingyun Liang and Jie Cao and Guolei Sun and K. Zhang and Luc Van Gool and Radu Timofte},
  journal={2021 IEEE/CVF International Conference on Computer Vision Workshops (ICCVW)},
  year={2021},
  pages={1833-1844},
  url={https://api.semanticscholar.org/CorpusID:237266491}
}

@article{Zamir2021RestormerET,
  title={Restormer: Efficient Transformer for High-Resolution Image Restoration},
  author={Syed Waqas Zamir and Aditya Arora and Salman Hameed Khan and Munawar Hayat and Fahad Shahbaz Khan and Ming-Hsuan Yang},
  journal={2022 IEEE/CVF Conference on Computer Vision and Pattern Recognition (CVPR)},
  year={2021},
  pages={5718-5729},
  url={https://api.semanticscholar.org/CorpusID:244346144}
}

@article{Jiang2017ANT,
  title={A Novel Tensor-Based Video Rain Streaks Removal Approach via Utilizing Discriminatively Intrinsic Priors},
  author={Tai-Xiang Jiang and Tingzhu Huang and Xile Zhao and Liang-Jian Deng and Yao Wang},
  journal={2017 IEEE Conference on Computer Vision and Pattern Recognition (CVPR)},
  year={2017},
  pages={2818-2827},
  url={https://api.semanticscholar.org/CorpusID:4910523}
}

@article{Ancuti2013SingleID,
  title={Single Image Dehazing by Multi-Scale Fusion},
  author={Codruta Orniana Ancuti and Cosmin Ancuți},
  journal={IEEE Transactions on Image Processing},
  year={2013},
  volume={22},
  pages={3271-3282},
  url={https://api.semanticscholar.org/CorpusID:14713713}
}

@article{Oppenheim1980TheIO,
  title={The importance of phase in signals},
  author={Alan V. Oppenheim and Jae Sung Lim},
  journal={Proceedings of the IEEE},
  year={1980},
  volume={69},
  pages={529-541},
  url={https://api.semanticscholar.org/CorpusID:23790114}
}

@inproceedings{Oppenheim1979PhaseIS,
  title={Phase in speech and pictures},
  author={Alan V. Oppenheim and Jae Sung Lim and Gary E. Kopec and Stephen C. Pohlig},
  booktitle={IEEE International Conference on Acoustics, Speech, and Signal Processing},
  year={1979},
  url={https://api.semanticscholar.org/CorpusID:23520948}
}

@inproceedings{Guo2022ExploringFP,
  title={Exploring Fourier Prior for Single Image Rain Removal},
  author={Xin Guo and Xueyang Fu and Man Zhou and Zhen Huang and Jialun Peng and Zhengjun Zha},
  booktitle={International Joint Conference on Artificial Intelligence},
  year={2022},
  url={https://api.semanticscholar.org/CorpusID:250638879}
}

@article{Wu2023QBenchAB,
  title={Q-Bench: A Benchmark for General-Purpose Foundation Models on Low-level Vision},
  author={Haoning Wu and Zicheng Zhang and Erli Zhang and Chaofeng Chen and Liang Liao and Annan Wang and Chunyi Li and Wenxiu Sun and Qiong Yan and Guangtao Zhai and Weisi Lin},
  journal={International Conference on Learning Representations},
  year={2024},
  volume={abs/2309.14181},
  url={https://api.semanticscholar.org/CorpusID:262824606}
}

@inproceedings{Luo2023ControllingVM,
  title={Controlling Vision-Language Models for Multi-Task Image Restoration},
  author={Ziwei Luo and Fredrik K. Gustafsson and Zheng Zhao and Jens Sjolund and Thomas B. Schon},
  booktitle={International Conference on Learning Representations},
  year={2023},
  url={https://api.semanticscholar.org/CorpusID:263605463}
}

@article{Chen2022LearningMA,
  title={Learning Multiple Adverse Weather Removal via Two-stage Knowledge Learning and Multi-contrastive Regularization: Toward a Unified Model},
  author={Wei-Ting Chen and Zhi-Kai Huang and Cheng-Che Tsai and Hao-Hsiang Yang and Jianping Ding and Sy-Yen Kuo},
  journal={2022 IEEE/CVF Conference on Computer Vision and Pattern Recognition (CVPR)},
  year={2022},
  pages={17632-17641},
  url={https://api.semanticscholar.org/CorpusID:250581068}
}

@article{Potlapalli2023PromptIRPF,
  title={PromptIR: Prompting for All-in-One Blind Image Restoration},
  author={Vaishnav Potlapalli and Syed Waqas Zamir and Salman Siddique Khan and Fahad Shahbaz Khan},
  journal={ArXiv},
  year={2023},
  volume={abs/2306.13090},
  url={https://api.semanticscholar.org/CorpusID:259224666}
}

@INPROCEEDINGS{10377637,
  author={Patil, Prashant W. and Gupta, Sunil and Rana, Santu and Venkatesh, Svetha and Murala, Subrahmanyam},
  booktitle={2023 IEEE/CVF International Conference on Computer Vision (ICCV)}, 
  title={Multi-weather Image Restoration via Domain Translation}, 
  year={2023},
  volume={},
  number={},
  pages={21639-21648},
  doi={10.1109/ICCV51070.2023.01983}}

@article{Liu2025FrequencyDD,
  title={Frequency Domain-Based Diffusion Model for Unpaired Image Dehazing},
  author={Chengxu Liu and Lu Qi and Jinshan Pan and Xueming Qian and Ming-Hsuan Yang},
  journal={2025 IEEE/CVF International Conference on Computer Vision (ICCV)},
  year={2025},
  pages={7538-7547},
  url={https://api.semanticscholar.org/CorpusID:280135104}
}

@inproceedings{Xu2024TowardsRA,
  title={Towards Real-World Adverse Weather Image Restoration: Enhancing Clearness and Semantics with Vision-Language Models},
  author={Jiaqi Xu and Mengyang Wu and Xiaowei Hu and Chi-Wing Fu and Qi Dou and Pheng-Ann Heng},
  booktitle={European Conference on Computer Vision},
  year={2024},
  url={https://api.semanticscholar.org/CorpusID:272368249}
}

@article{Wang2025LearningHT,
  title={Learning Hazing to Dehazing: Towards Realistic Haze Generation for Real-World Image Dehazing},
  author={Ruiyi Wang and Yushuo Zheng and Zicheng Zhang and Chunyi Li and Shuaicheng Liu and Guangtao Zhai and Xiaohong Liu},
  journal={2025 IEEE/CVF Conference on Computer Vision and Pattern Recognition (CVPR)},
  year={2025},
  pages={23091-23100},
  url={https://api.semanticscholar.org/CorpusID:277313707}
}

@misc{tian2025yolov12attentioncentricrealtimeobject,
      title={YOLOv12: Attention-Centric Real-Time Object Detectors}, 
      author={Yunjie Tian and Qixiang Ye and David Doermann},
      year={2025},
      eprint={2502.12524},
      archivePrefix={arXiv},
      primaryClass={cs.CV},
      url={https://arxiv.org/abs/2502.12524}, 
}

@article{Fu_2017,
   title={Clearing the Skies: A Deep Network Architecture for Single-Image Rain Removal},
   volume={26},
   ISSN={1941-0042},
   url={http://dx.doi.org/10.1109/TIP.2017.2691802},
   DOI={10.1109/tip.2017.2691802},
   number={6},
   journal={IEEE Transactions on Image Processing},
   publisher={Institute of Electrical and Electronics Engineers (IEEE)},
   author={Fu, Xueyang and Huang, Jiabin and Ding, Xinghao and Liao, Yinghao and Paisley, John},
   year={2017},
   month=June, pages={2944–2956} }

@misc{
sharif2026degradationaware,
title={Degradation-Aware All-in-One Image Restoration via Latent Prior Encoding},
author={S M A Sharif and Abdur Rehman and Fayaz Ali and Radu Timofte and Rizwan Ali Naqvi},
year={2026},
url={https://openreview.net/forum?id=4fhbxglvYs}
}

@inproceedings{Wu2025GradientAC,
  title={Gradient as Conditions: Rethinking HOG for All-in-one Image Restoration},
  author={Jiawei Wu and Zhifei Yang and Zhe Wang and Zhi Jin},
  booktitle={AAAI Conference on Artificial Intelligence},
  year={2026},
  url={https://api.semanticscholar.org/CorpusID:277781355}
}

@article{Wang2025MODEMAM,
  title={MODEM: A Morton-Order Degradation Estimation Mechanism for Adverse Weather Image Recovery},
  author={Hainuo Wang and Qiming Hu and Xiaojie Guo},
  journal={Conference on Neural Information Processing Systems},
  year={2025},
  volume={abs/2505.17581},
  url={https://api.semanticscholar.org/CorpusID:278886994}
}

@article{Huang2023ContrastiveSL,
  title={Contrastive Semi-Supervised Learning for Underwater Image Restoration via Reliable Bank},
  author={Shirui Huang and Keyan Wang and Huan Liu and Jun Chen and Yunsong Li},
  journal={2023 IEEE/CVF Conference on Computer Vision and Pattern Recognition (CVPR)},
  year={2023},
  pages={18145-18155},
  url={https://api.semanticscholar.org/CorpusID:257557240}
}
\bibliographystyle{iclr2027_conference}

\newpage
\appendix

\section{Ablation studies}

\subsection{Spectral visualizations}
\label{app:spectral}
\begin{figure}[!t]
    \centering
    \includegraphics[width=0.8\linewidth]{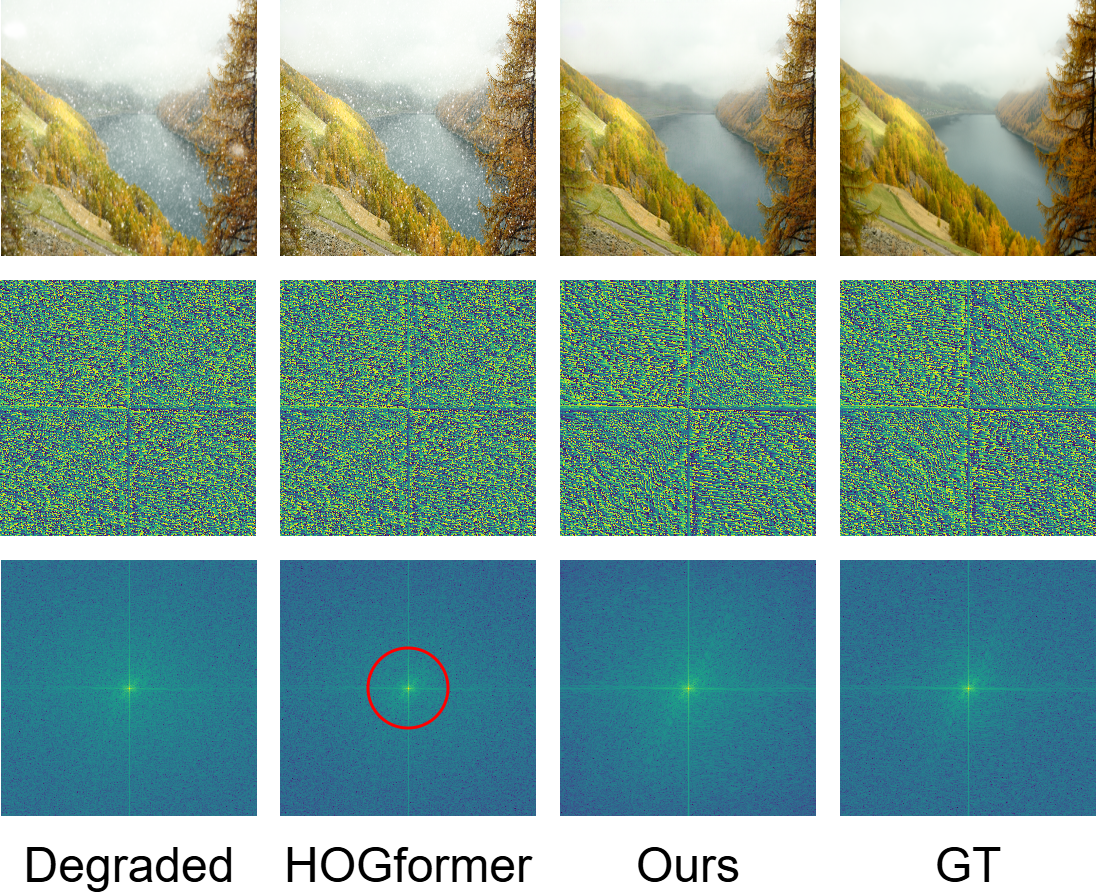}
    \caption{Visual results on CDD11 dataset. From top to bottom: images, phase spectra, and amplitude spectra. Zoom in to see better visualization.}
    \label{fig:desnow_spectral}
\end{figure}

To prove the reliability of our model,
we present the spectral visualizations in Fig.\ \ref{fig:desnow_spectral} on CDD11 dataset. Our model's results are more aligned with the ground truth than results from HOGformer. In addition, we measure the L2 distance ($\downarrow$) of the frequency features with GT to show the superiority and reported in Tab.\ \ref{tab:l2_distance}. Compared with HOGFormer, our method substantially reduces both phase and amplitude errors across all three degradation types. The largest improvements are observed for rain and haze, where the phase error decreases from 52.645 to 33.735 and 57.057 to 32.623, respectively. Amplitude error is also consistently reduced, with the average decreasing from 11.184 to 8.111. Overall, our method lowers the average phase error from 57.453 to 41.667, demonstrating better preservation and recovery of both structural phase information and frequency amplitude characteristics across different degradations.



\begin{table}[!h]
\centering
\renewcommand{\arraystretch}{1.15}
\begin{tabular}{c|cccc}
\hline
\multirow{3}{*}{\begin{tabular}[c]{@{}c@{}}Degradation \\ type\end{tabular}} & \multicolumn{4}{c}{Methods}                                        \\ \cline{2-5} 
                                                                             & \multicolumn{2}{c|}{HOGFormer}          & \multicolumn{2}{c}{Ours} \\ 
                                                                             & Phase  & \multicolumn{1}{c|}{Amplitude} & Phase     & Amplitude    \\ \hline
Rain                                                                         & 52.645 & \multicolumn{1}{c|}{12.331}    &\textbf{ 33.735}    & \textbf{8.836}        \\ \hline
Snow                                                                         & 62.657 & \multicolumn{1}{c|}{9.616}     & \textbf{58.644}    & \textbf{8.953}       \\ \hline
Haze                                                                         & 57.057 & \multicolumn{1}{c|}{11.604}    & \textbf{32.623}    & \textbf{6.545}        \\ \hline
Avg                                                                          & 57.453 & \multicolumn{1}{c|}{11.184}    & \textbf{41.667}    & \textbf{8.111}       
\end{tabular}
\caption{Average L2$\downarrow$ distance of the frequency features with GT on CDD11 dataset.}
\label{tab:l2_distance}
\end{table}

\subsection{Overhead comparison with other SOTAs}
\label{app:overhead_sota}
We have conducted an overhead comparison with HOGformer \citep{Wu2025GradientAC} and MODEM \citep{Wang2025MODEMAM} with the input of $256\times256$ on a single H100 GPU. The results, shown in Tab.\ \ref{tab:overhead_comparison}, indicate that our model provides a favorable efficiency profile compared with existing approaches. While using a comparable number of trainable parameters to MODEM and HOGFormer, it achieves substantially lower inference latency at only 21.20 ms, compared with 100.83 ms and 218.82 ms, respectively. It also requires the least GPU memory (497.08 MB), despite having a larger overall model size of 225 MB. These results indicate that our method is particularly efficient at inference time, offering faster processing and lower runtime memory consumption while maintaining competitive parameter complexity.
\begin{table}[!h]
\centering
\renewcommand{\arraystretch}{1.15}
\begin{tabular}{l|c|c|c|c}
\hline
\multirow{2}{*}{Method} & \begin{tabular}[c]{@{}c@{}}Trainable\\ Param\end{tabular} & Latency & GPU  Mem & Model Size \\
                        & M                                                         & ms      & MB       & MB         \\ \hline
MODEM                   & 19.96                                                     & 100.83  & 551.21   & 77         \\
HOGformer               & 16.65                                                     & 218.82  & 760.43   & 65         \\ \hline
Ours                    & 19.67                                                     & 21.20   & 497.08   & 225       
\end{tabular}
\caption{Inference overhead comparison with other SOTAs. Input size of $256\times256$ on a single H100.}
\label{tab:overhead_comparison}
\end{table}

\subsection{Overhead comparison with other unsupervised constraint methods}
\label{app:overhead_unsupervised_constraints}
We have conducted an overhead comparison with LPIPS and text (using VLM) with the input of $256\times256$ on a single A100 GPU. In this experiment, we want to capture the GPU usage and latency that models take to produce a semantic signal from the input image. The results, shown in Tab.\ \ref{tab:overhead_comparison}, indicate that our model achieves the best tradeoff of inference time and performance. The average performance of three tasks (deraining, desnowing, and dehanzing) is also cited from the main paper. Compared with the text-based constraint, the phase-based approach reduces latency from 1198.14 ms to 41.00 ms and GPU memory consumption from 13,846.31 MB to 2,894.09 MB, while achieving better performance on both LIQE (3.527) and CLIP-IQA (0.572). Although LPIPS introduces the lowest computational overhead, its average LIQE score is lower. Overall, the phase-based constraint achieves the strongest perceptual performance while remaining substantially more efficient than the text-based alternative.

\begin{table}[!t]
\centering
\renewcommand{\arraystretch}{1.15}
\begin{tabular}{l|c|c|cc}
\hline
\multirow{2}{*}{Method} & Latency & GPU mem  & \multicolumn{2}{c}{Average Performance} \\
                        & ms      & MB       & LIQE$\uparrow$              & CLIP-IQA$\uparrow$            \\ \hline
Using LPIPS             & 2.14    & 1774.66  & 3.368             & 0.565               \\
Using Text              & 1198.14 & 13846.31 & 3.407             & 0.545               \\ \hline
Using Phase             & 41.00   & 2894.09  & 3.527             & 0.572              
\end{tabular}
\caption{Overhead comparison with other constraint methods. Input size of $256\times256$ on a single A100.}
\end{table}
\subsection{Impact of different backbones}
\label{acc:backbones}
We evaluate our proposed methods with different backbone models and report the results in Tab.\ \ref{tab:backbone}. The comparison across different backbones shows that the proposed framework can be effectively integrated with multiple restoration architectures. Among them, NAFNet achieves the best perceptual performance, reaching 3.545 LIQE and 0.574 CLIP-IQA, but at the cost of substantially higher computational complexity with 67.89M trainable parameters and 290.84G FLOPs. Uformer is considerably lighter but yields lower restoration quality. MSBDN provides a more balanced configuration, achieving 3.527 LIQE and 0.572 CLIP-IQA with fewer parameters and FLOPs than NAFNet. These results indicate that the choice of backbone allows a flexible trade-off between computational cost and perceptual restoration performance.

\begin{table}[]
\centering
\renewcommand{\arraystretch}{1.15}
\begin{tabular}{l|c|c|cc}
\hline
\multirow{2}{*}{Backbone} & \begin{tabular}[c]{@{}c@{}}Trainable\\ Param\end{tabular} & FLOP   & \multicolumn{2}{c}{Average Performance} \\
                        & M                                                         & G      & LIQE$\uparrow$               & CLIP-IQA$\uparrow$             \\ \hline
NAFNet                  & 67.89                                                     &   290.84     &    \textbf{3.545}                &        \textbf{0.574}            \\
Uformer                 & 5.29                                                      &  42.74      &            3.268           &  0.497                  \\
MSBDN                   & 19.67                                                     & 153.94 & \underline{3.527}    & \underline{0.572}  
\end{tabular}
\caption{Results of different backbone models. We report the average performance on three tasks using the dataset presented in the main paper.}
\label{tab:backbone}
\end{table}
\subsection{Abalation study on adaptive weight $w$} \label{app:adaptive_weight}

\begin{figure}[!t]
    \centering
    \includegraphics[width=0.5\linewidth]{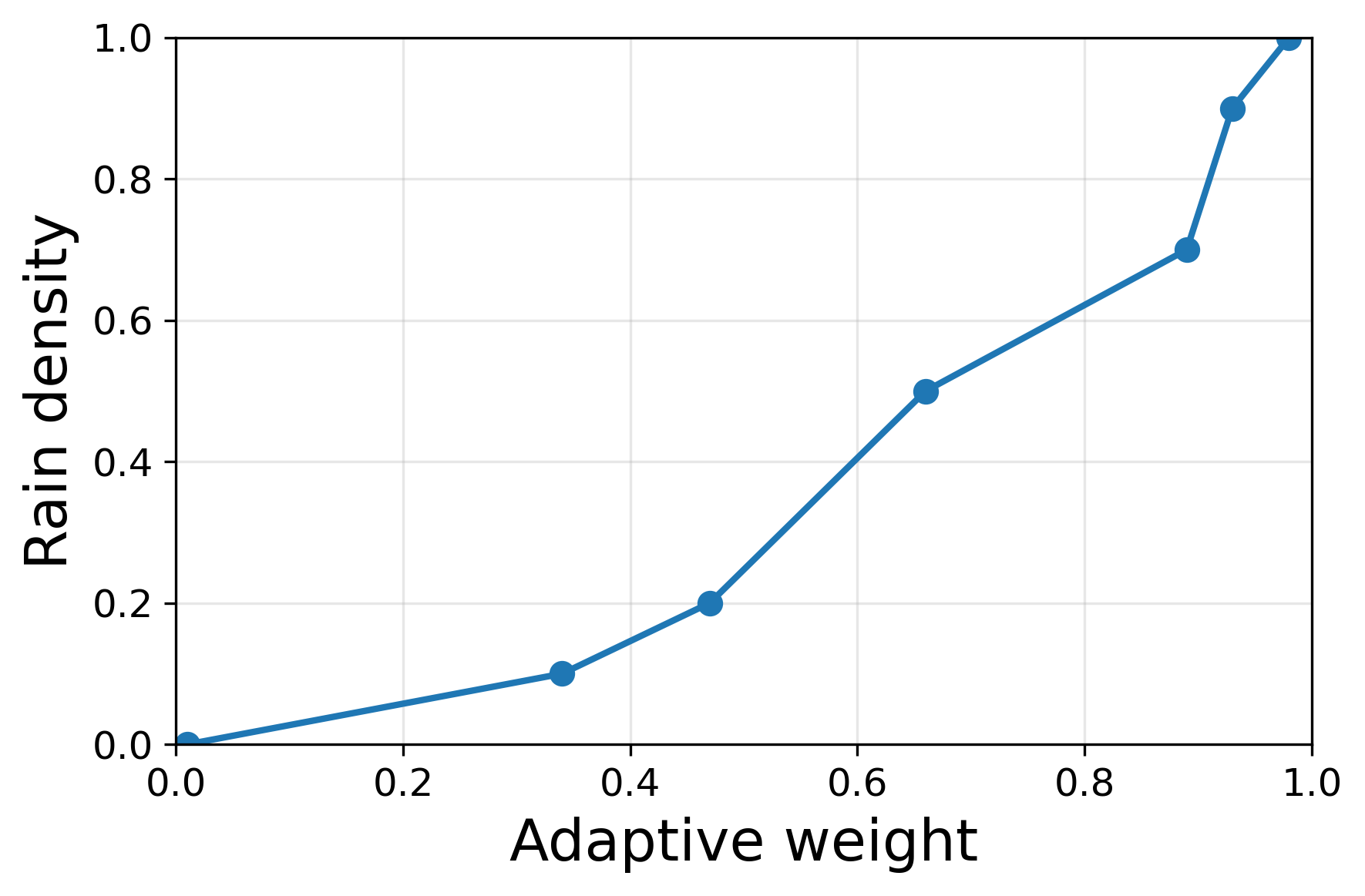}
    \caption{Relationship between adaptive weight and rain density. Higher rain densities are assigned larger adaptive weights, enabling the model to place greater emphasis on pseudo-GT.}
    \label{fig:w_vs_density}
\end{figure}

Figure \ref{fig:w_vs_density} illustrates how the adaptive weight varies with rain density. The weight increases monotonically as the degradation becomes more severe. For light rain, the weight remains relatively low, while a sharper increase is observed at higher densities, particularly beyond a density of 0.7. This behavior allows the model to adapt its restoration strength according to degradation severity. In this experiment, we use synthetic rain so that we can control the rain density.
\subsection{Impact of unreliable data}
\label{app:unreliable}
\begin{figure}[!t]
        \centering
        \includegraphics[width=0.8\linewidth]{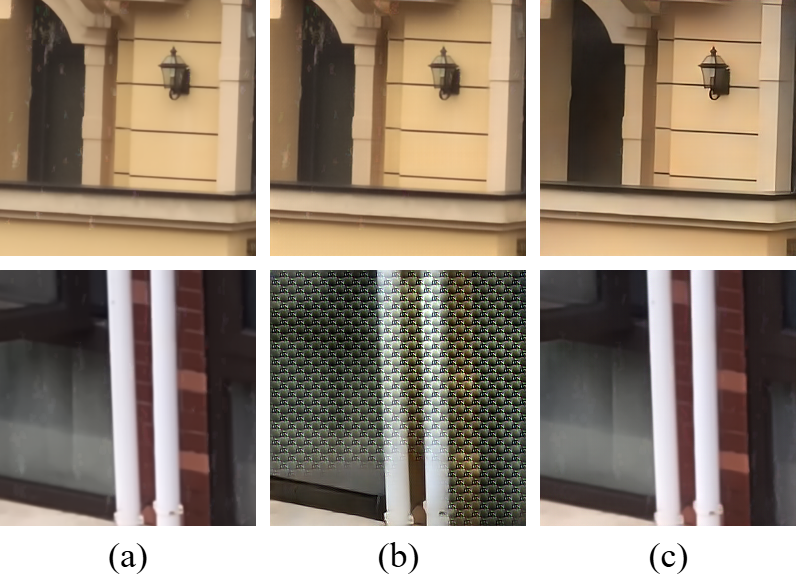}
        \caption{(a) Input image. Output of teacher model: (b) without unreliable database; (c) with unreliable database.}
        \label{fig:test_with_teacher}
    \end{figure}

We further validate the effectiveness of the unreliable database in Fig.~\ref{fig:test_with_teacher}. Compared with the baseline, the teacher network produces more reliable pseudo-GTs with fewer artifacts and distortions, resulting in improved supervision for the student.

\section{More Experimental Results}
In this section, we first provide results on downstream tasks and more quantitative results. Then, we show more qualitative results. All of these experiments are conducted in the all-in-one setting.

\subsection{Downstream task results - Object Detection (OD)} \label{app:od_task}
To evaluate our model on downstream tasks, we follow the setting of \citep{sharif2026degradationaware}. We construct a comprehensive evaluation framework using 1,000 test images spanning three common image degradations: deraining \citep{Fu_2017}, dehazing \citep{Li2017BenchmarkingSD}, and desnowing \citep{Liu2017DesnowNetCD}. Leveraging the paired nature of these datasets, we first annotate the clean ground-truth images and transfer the annotations to their corresponding degraded counterparts, ensuring consistent and reliable labels. We then restore the degraded images using several baseline all-in-one image restoration methods and evaluate their downstream object detection performance using YOLOv12-Large \citep{tian2025yolov12attentioncentricrealtimeobject}.

\begin{table}[]
\centering
\begin{tabular}{l|ccc|ccc}
\hline
\multirow{2}{*}{Method} & \multicolumn{3}{c|}{AP}           & \multicolumn{3}{c}{AP by Object Size} \\
                        & IoU=0.5:0.95 & IoU=0.5 & IoU=0.75 & Small      & Medium      & Large      \\ \hline
Uformer                 & 29.6         & 31.3    & 29.9     & 37.5       & 27.3        & 27.9       \\
AirNet                  & 31.0         & 32.9    & 32.1     & 36.1       & 26.6        & 30.7       \\
PromptIR                & 31.2         & 33.7    & 32.3     & 42.1       & \textbf{29.9}        & 31.1       \\
DFPIR                   & 28.8         & 30.9    & 29.8     & 27.4       & 26.4        & 30.1       \\ \hline
Ours                    & \textbf{35.3}         & \textbf{38.7}    & \textbf{36.6 }    & \textbf{44.4}       & \textbf{29.9}        & \textbf{35.3 }      \\ \hline
\end{tabular}
\caption{Performance comparison of object detection models. We report average precision (AP) across different IoU thresholds and object sizes.}

\label{tab:downstream_task}
\end{table}

Tab.~\ref{tab:downstream_task} shows that our method achieves the strongest overall performance across all evaluation metrics. It improves the overall AP to 35.3, outperforming the best baseline, PromptIR \cite{Potlapalli2023PromptIRPF}, by 4.1 points. Similar gains are observed at different IoU thresholds, reaching 38.7 AP\(_{50}\) and 36.6 AP\(_{75}\), which indicates improvements in both detection accuracy and localization quality. For object-scale evaluation, our method obtains the highest performance on small (44.4) and large (35.3) objects, while matching the best result on medium-sized objects (29.9). These results demonstrate that the proposed method provides consistent improvements across different IoU criteria and object scales.
\begin{table}[b]
\centering
\renewcommand{\arraystretch}{1.15}
\begin{tabular}{c|c|cc}
\hline
\multirow{2}{*}{\begin{tabular}[c]{@{}c@{}}Degradation\\ types\end{tabular}} & \multirow{2}{*}{Metrics} & \multicolumn{2}{c}{Methods}       \\
                                  &                          & HOGformer       & Ours            \\ \hline
\multirow{2}{*}{Rain}             & Q-Align$\uparrow$                 & 2.8022          & \textbf{2.8372} \\
                                  & MUSIQ$\uparrow$                     & 53.958          & \textbf{55.776} \\ \hline
\multirow{2}{*}{Snow}             & Q-Align$\uparrow$                  & \textbf{2.7557} & 2.6678          \\
                                  & MUSIQ$\uparrow$                    & \textbf{54.947} & 52.336          \\ \hline
\multirow{2}{*}{Haze}             & Q-Align$\uparrow$                  & 2.4597          & \textbf{2.5241} \\
                                  & MUSIQ$\uparrow$                    & 50.438          & \textbf{51.710} \\ \hline
\multirow{2}{*}{Average}          & Q-Align$\uparrow$                   & 2.6725          & \textbf{2.6764} \\
                                  & MUSIQ$\uparrow$                    & 53.114          & \textbf{53.274}
\end{tabular}
\caption{Quantitative results on different metrics.}
\label{tab:qalign_musiq}
\end{table}
\subsection{More quantitative results with different metrics}
\label{app:diff_metrics}
To further prove the effectiveness of our all-in-one model, we have conducted a comparison with HOGformer \citep{Wu2025GradientAC} in two different no-reference metrics, Q-Align \citep{wu2023qalign} and MUSIQ \citep{Ke2021MUSIQMI}. The results, which are reported in Tab.\ \ref{tab:qalign_musiq}, indicate that our method achieves the best average performance across both perceptual quality metrics, improving Q-Align from 2.6725 to 2.6764 and MUSIQ from 53.114 to 53.274. In particular, our method consistently outperforms HOGFormer on both metrics for rain degradation, while also providing clear improvements for haze restoration. For snow degradation, HOGFormer performs better, suggesting that our method may be less effective for this specific corruption type. Nevertheless, the higher overall average indicates that our approach provides more balanced restoration quality across different degradation conditions. We use the same dataset as the main paper.










\end{document}